\documentclass[letterpaper, 10 pt, conference]{ieeeconf}  % Comment this line out if you need a4paper

\IEEEoverridecommandlockouts                              % This command is only needed if 
\usepackage{graphicx}
\usepackage{subcaption}
\usepackage{amsmath} % assumes amsmath package installed
\usepackage{amssymb}  % assumes amsmath package installed

\title{\LARGE \bf
Re-evaluating Parallel Finger-tip Tactile Sensing \\for Inferring Object Adjectives: An Empirical Study
}

\author{Fangyi Zhang$^{1}$, Peter Corke $^{1}$%
\thanks{$^{1}$Authors are with the Queensland University of Technology (QUT) Centre for Robotics, 2 George Street, Brisbane City, 4000, Queensland, Australia. (email: fangyi.zhang@qut.edu.au; peter.corke@qut.edu.au)}
}

\begin{document}

\maketitle
\thispagestyle{empty}
\pagestyle{empty}

\begin{abstract}

Finger-tip tactile sensors are increasingly used for robotic sensing to establish stable grasps and to infer object properties. Promising performance has been shown in a number of works for inferring adjectives that describe the object, but there remains a question about how each taxel contributes to the performance. This paper explores this question with empirical experiments, leading insights for future finger-tip tactile sensor usage and design.

\end{abstract}

\section{Introduction}
In recent years, with the developments of tactile sensors and machine learning techniques, finger-tip tactile sensors are increasingly used for robotic tasks such as adjective classification~\cite{chu2015robotic}. Promising performance has been shown for adjective classification~\cite{chu2015robotic,gao2016deep,richardson2019improving} or learning adjective distributions~\cite{richardson2020learning}. However, there are few works exploring: \textit{how each taxel in a finger-tip tactile sensor contributes to the inferring of object adjectives}, and \textit{how taxel spatial density and measurement temporal density influence the performance}. This paper seeks answers to these questions via empirical experiments.

\begin{figure}[t]
    \centering
    \includegraphics[width=0.7\linewidth]{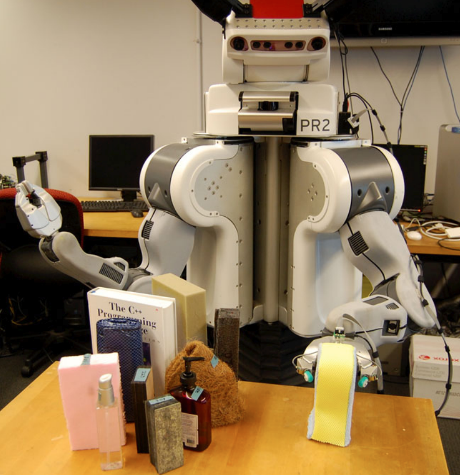}
    \caption{A PR2 equipped with BioTac sensors explores an object in this case a blue sponge, for collecting the tactile dataset Penn Haptic Adjective Corpus (PHAC-2)~\cite{chu2015robotic}.}
    \label{fig:dataset_collection}
\end{figure}

The empirical study is conducted using the PHAC-2 dataset~\cite{chu2015robotic}. 
The dataset was collected using a Willow Garage PR2 equipped with a pair of SynTouch BioTac sensors to explore a set of objects using a set of finger-tip motions (see Fig.~\ref{fig:dataset_collection}).
To focus on the benefit of tactile signals, rather than measuring classification performance, we propose to evaluate the information quality of the tactile signals, i.e., measurements indicating how tactile signals could potentially perform in downstream tasks such as the adjective classification task, but avoiding the influence of classifiers which might introduce differences or biases due to method selection or parameter tuning. 
Detailed method design is introduced in Section~\ref{sec:meth}, followed by experiments (Section~\ref{sec:exp_results}) 
seeking answers to the aforementioned questions. This paper makes three major contributions:
\begin{itemize}
    \item a new method is proposed to re-evaluate finger-tip tactile data for inferring object adjectives from an information quality rather than classification performance perspective,
    \item comprehensive experiments are conducted to study the benefit of finger-tip tactile sensing from the perspectives of taxel distribution, taxel spatial density and measurement temporal density, and
    \item important insights are obtained to guide the usage and design of finger-tip tactile sensors in related tasks.
\end{itemize}

\section{Related Works}

There are a number of works exploring tactile sensing for inferring object adjectives~\cite{liu2017recent,navarro2022visuo}, many of which use the PHAC-2 dataset~\cite{chu2015robotic}. Early solutions use hand-crafted features together with an SVM for adjective classification~\cite{chu2015robotic}, achieving promising performance. Better performance was then achieved by various learning approaches~\cite{liu2016extreme,liu2017multi,liu2017structured}. Approaches were also proposed to fuse vision and tactile for adjective classification, showing that the fusion of the two modalities improves the performance~\cite{gao2016deep,abderrahmane2018visuo}. The current State-of-The-Art classification performance was achieved by a method using learned features~\cite{richardson2019improving}. More recently, a learning method was proposed to predict perceptual distributions of haptic adjectives~\cite{richardson2020learning}, demonstrating the feasibility of modeling both the intensity and the variation of tactile perception. All these methods enabled better adjective classification or perceptual distribution prediction. Some of them also studied how different types of signals and different exploratory procedures contributed to adjective classification. However, no study has been done at the taxel level. It remains unknown how much each taxel contributes to inferring the object adjectives.

\section{Methodology}
\label{sec:meth}

The exploration of how the finger-tip tactile information contributes to a robotic task is conducted by ablating the tactile signals and evaluating information quality measures such as distinguishability which is critical for downstream tasks such as the adjective classification~\cite{chu2015robotic}. 
Signals with a higher distinguishability have a more distinct distribution in their feature space, and are therefore better for inferring object adjectives (e.g., easier to classify).
The tactile signals are ablated from the perspectives of \textbf{taxel distribution}, \textbf{taxel spatial density}, and \textbf{measurement temporal density}.

\subsection{Dataset}
\label{sec:dataset}
The study is conducted using the tactile adjective classification dataset Penn Haptic Adjective Corpus (PHAC-2)~\cite{chu2015robotic}. 
The dataset contains 600 trials collected using a Willow Garage PR2 equipped with two SynTouch BioTac tactile finger-tip sensors as shown in Fig.~\ref{fig:dataset_collection}. 
The trials were recorded by using the robot to perform four exploratory procedures (EPs): ``squeeze'', ``hold'', ``slow slide'', and ``fast slide'' on 60 different objects, with 10 trials of each. 
A tactile signal frame $\mathbf{V}$ contains signals from both left ($\mathbf{V}_{L}$) and right ($\mathbf{V}_{R}$) BioTac sensors, each of which comprises 19 spatially distributed electrodes (taxels) $\mathbf{E} \in \mathbb{R}^{19}$ and four scalars: vibration ($P_{AC}$), pressure ($P_{DC}$), heat flow ($T_{AC}$) and temperature ($T_{DC}$), i.e., $\mathbf{V}_{L}, \mathbf{V}_{R} \in \mathbb{R}^{23}$.
The $P_{AC}$ is measured at 2.2 kHz, while the other signals are measured at 100 Hz.

The dataset was designed for adjective classification, but in this work it is used for tactile signal ablation and information quality evaluation in this work. 
The dataset has 19 binary adjective labels $\mathbf{A} \in \mathbb{B}^{19}$ (the version~\cite{chu2015robotic} has 24 adjectives, but we follow the setup in~\cite{richardson2019improving} in this work and exclude those with fewer than three positively labeled objects). We use the same training and test sets as~\cite{chu2015robotic} and~\cite{richardson2019improving}.

\subsection{Metrics}
\label{sec:metrics}
The information quality (distinguishability) of tactile signals are evaluated using the commonly used mean Average Precision (\textbf{mAP})~\cite{zhu2004recall} which directly indicates how signals could potentially perform in downstream tasks such as classification, but with no influence from classifiers which might introduce differences or biases due to method selection or parameter tuning.
\textbf{mAP} is a good indication of how samples of different classes are distinctly distributed in the feature space. A larger \textbf{mAP} means a more distinct distribution of samples across different classes, and is therefore easier to classify.
When comparing to existing classification baselines such as~\cite{chu2015robotic} and~\cite{richardson2019improving}, $\mathbf{F_1}$ score~\cite{taha2015metrics} is used to measure classification performance in Section~\ref{sec:baseline_results}. 

\subsection{Feature generation}
\label{sec:meth_pca}
Different from~\cite{chu2015robotic} and~\cite{richardson2019improving} where hand-crafted or learned features were used, dimension-reduced raw tactile signals are directly used as features for information quality evaluation in this work to avoid unnecessary biases. The dimension reduction is performed via Principal Component Analysis (PCA)~\cite{jolliffe2016principal}.

A sequence of tactile signals from one sensor from one exploratory procedure 
are first concatenated into one vector $\mathbf{Z}_k = [\mathbf{V}_{k,1}, \mathbf{V}_{k,2}, \cdots, \mathbf{V}_{k,N}] \in \mathbb{R}^{23N}, k \in [L,R]$,
where $N$ is the total number of frames in one sequence.
Then $\mathbf{Z}_{L}$ and $\mathbf{Z}_{R}$ from the two fingers are concatenated to $\mathbf{Z} \in \mathbb{R}^{46I}$. PCA is then applied to $\mathbf{Z} \in \mathbb{R}^{46I}$. The top 200 components are used as the features representing each sequence for information quality evaluation. Evaluation experiments in Section~\ref{sec:experiment_pca} show that the top 200 components could maintain $>99\%$ of the input's variance. For the convenience of expression, the dimension-reduced features are denoted as $\mathbf{Z}^{200} \in \mathbb{R}^{200}$. 

Table~\ref{tab:length_of_each_EP} shows the $N$ values and feature dimensions for the tactile sequences collected from the four exploratory procedures. The $N$ values we used are the maximum length of the sequences for each EP. The sequences with a shorter length are padded to the maximum length by repeating their first and last elements in a symmetric manner (the non-repeated fractions are in the middle of the padded signals). 

\begin{table}[]
    \centering
    \begin{tabular}{c|c|c|c}
    \hline
         EP & $N$ & Dimension of $\mathbf{Z}$ & Dimension reduced (\%) \\
         \hline
         Squeeze & 4740 & 218,040 & 99.9\\
         Hold & 2772 & 127,512 & 99.8\\
         Slow Slide & 1784 & 82,064 & 99.8\\
         Fast Slide & 1714 & 78,844 & 99.7\\
        \hline
    \end{tabular}
    \caption{The total number of frames $N$ used for each EP and feature dimensions.}
    \vspace{-1em}
    \label{tab:length_of_each_EP}
\end{table}

\subsection{Classification method}
\label{sec:meth_svm}
Although the main focus of this work is to explore the benefit of each tactile signal by ablating raw signals and observing their information quality changes, classification performance is also evaluated on the dimension-reduced features $\mathbf{Z}^{200}$ to see their feasibility by comparing to the initial method~\cite{chu2015robotic} (SVM with hand-crafted features, denoted as \textbf{HF+SVM}) and the current State-of-The-Art solution~\cite{richardson2019improving} (SVM with learned features, denoted as \textbf{LF+SVM}) for adjective classification with PHAC-2. 

Similarly to~\cite{chu2015robotic,richardson2019improving}, SVM is adopted to predict binary adjective labels. Nineteen SVMs are trained for each EP (one for each adjective). Cross-validation and grid search are used to select the SVM regularization parameter $C$ with values ranging from $10^{-4}$ to $10^4$, tolerance $tol$ for the stopping criterion with values ranging from $10^{-6}$ to $10$, and the kernel either linear or RBF~\cite{kecman2005support}.
Similar to the settings in~\cite{chu2015robotic}, each classifier is trained using the $L_2$ metric~\cite{burges1998tutorial} with the $\mathbf{F_1}$ score averaged over 100 cross-validation sets created by randomly selecting 3 objects from each training set.

\section{Experimental Results}
\label{sec:exp_results}
Experiments are first conducted to see the feasibility of the dimension-reduced features by evaluating their classification performance with an SVM as introduced in Section~\ref{sec:meth_svm}.
This is followed by ablation studies to show the benefits of PCA. Then experiments are conducted to study how the individual tactile signals contribute to or influence the adjective inferring performance.

\subsection{Classification performance}
\label{sec:baseline_results}

\begin{figure}[!t]
      \centering
      \includegraphics[width=1.0\linewidth]{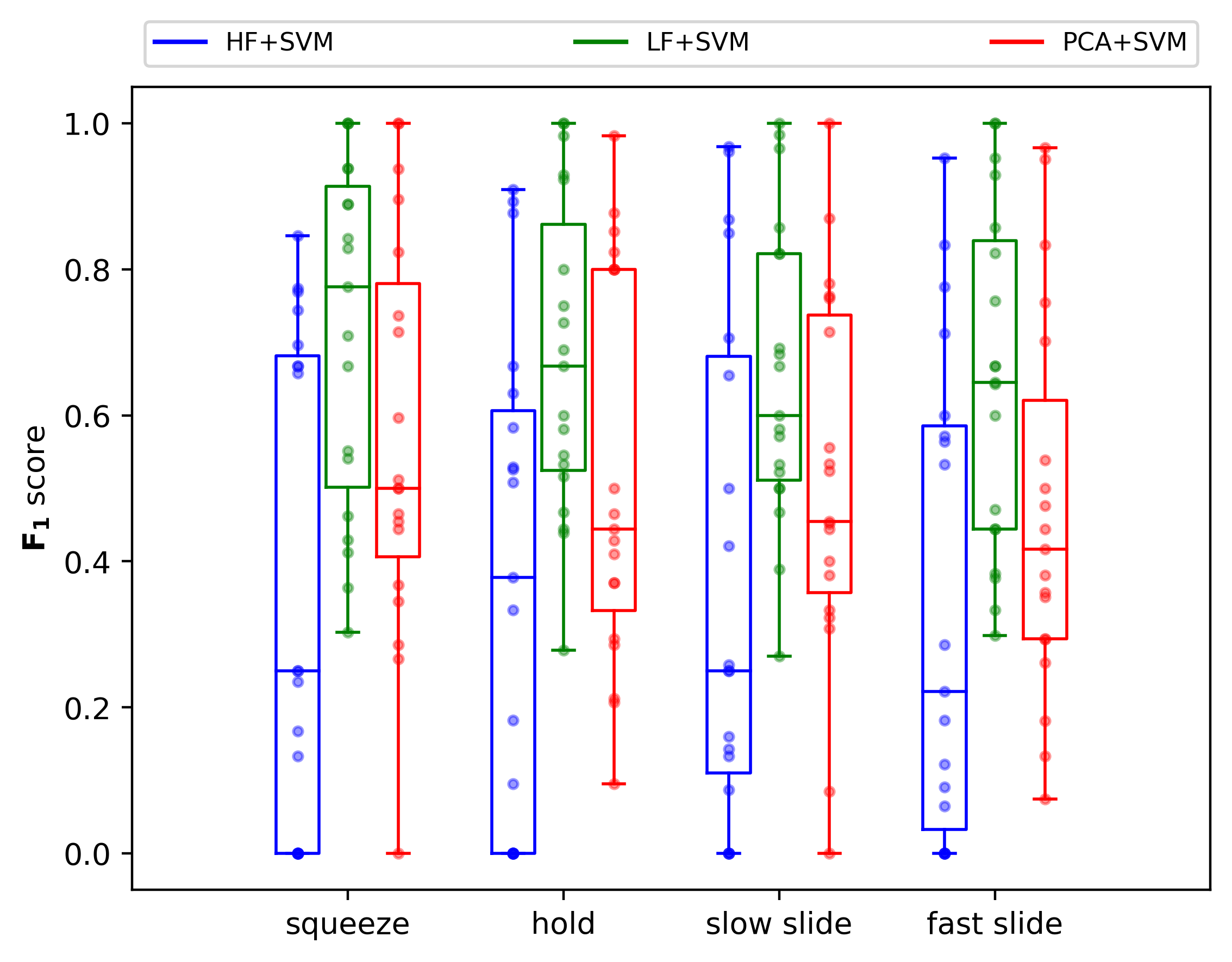}
      \caption{Boxplots of $\mathbf{F_1}$ scores of \textbf{HF+SVM}, \textbf{LF+SVM} and \textbf{PCA+SVM}. Each box shows the results of one method for the 19 adjectives w.r.t. EPs.}
      \label{fig:baseline_comp_boxplt}
    \end{figure}
  
    \begin{figure}[!t]
      \centering
      \includegraphics[width=1\linewidth]{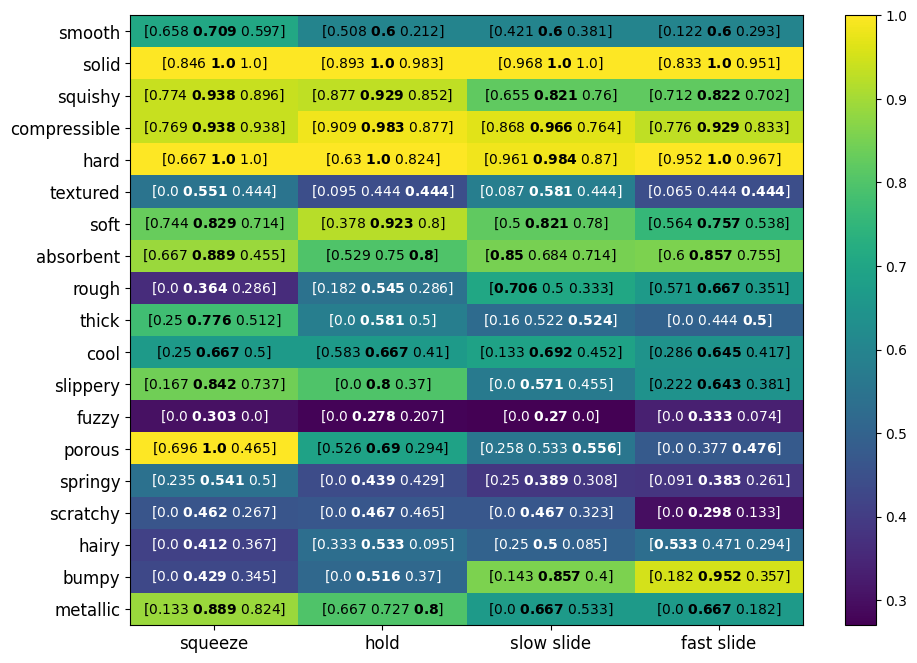}
      \caption{$\mathbf{F_1}$ scores of the three classification methods. Their values are shown as ``$[$ \textbf{HF+SVM} \textbf{LF+SVM} \textbf{PCA+SVM}$]$'', with the best ones bolded and their values indicated by the grid colors.}
      \label{fig:baseline_comp_color_table}
    \end{figure}

The classification performance of SVM with the dimension-reduced features (denoted as \textbf{PCA+SVM}) was evaluated by comparing to \textbf{HF+SVM} and \textbf{LF+SVM} using $\mathbf{F_1}$ scores. \textbf{mAP} 
was not compared as there were no \textbf{mAP} results presented in~\cite{chu2015robotic} and~\cite{richardson2019improving}. The setups for the dataset, PCA and SVM were introduced in Section~\ref{sec:meth}.

\begin{table}[!t]
    \centering
    \begin{tabular}{c|c|c|c|c}
    \hline
         & Squeeze & Hold & Slow Slide & Fast Slide \\
    \hline
        \textbf{HF+SVM} & 0.361 & 0.374 & 0.379 & 0.343\\
        \textbf{LF+SVM} & 0.713 & 0.677 & 0.654 & 0.647\\
        \textbf{PCA+SVM} & 0.571 & 0.527 & 0.510 & 0.469\\
    \hline
    \end{tabular}
    \caption{Averge $\mathbf{F_1}$ scores of different methods w.r.t. EPs.}
    \vspace{-1em}
    \label{tab:baseline_avg_f1_scores}
\end{table}

Fig.~\ref{fig:baseline_comp_boxplt} shows the boxplots of the $\mathbf{F_1}$ scores of the three methods. We can see that \textbf{PCA+SVM} performs better than \textbf{HF+SVM}, but worse than \textbf{LF+SVM} across all EPs. Considering that the three works all use SVM-based classifiers with similar settings, this result indicates that the features produced by the proposed PCA-based dimension reduction method works better than the hand-crafted features, but worse than the learned features.

Detailed $\mathbf{F_1}$ scores are shown in Fig.~\ref{fig:baseline_comp_color_table}, from which we can see that \textbf{PCA+SVM} has a performance in between in most cases (42/76), 
with 54 cases better than \textbf{HF+SVM} and 12 cases being the best.
This is also indicated by the average $\mathbf{F_1}$ scores listed in Table~\ref{tab:baseline_avg_f1_scores} where \textbf{PCA+SVM} has an in-between average score across all EPs.
These results demonstrate the feasibility of the dimension-reduced features generated directly from the raw tactile signals via PCA. 

Similarly to~\cite{chu2015robotic} and \cite{richardson2019improving}, we can also observe from Fig.~\ref{fig:baseline_comp_color_table} that the tactile signals collected by the four EPs are not informative enough for some texture-related adjectives (whose best $\mathbf{F_1}$ scores are below 0.75) such as: ``smooth'', ``textured'', ``rough'', ``fuzzy'', ``scratchy'', and ``hairy''. As mentioned in \cite{chu2015robotic}, this poor performance might be related to the degradation of the ridges on the BioTac sensor skin during data collection. The best scores of ``cool'' and ``springy'' are also below 0.75, which might also be related to similar issues (not much useful information was obtained in data collection). From the EP perspective, ``squeeze'' is the most informative ($\mathbf{F_1}$ scores above 0.75 in \textbf{10} adjectives); then ``hold'' comes second (performs good in \textbf{8} adjectives); the two sliding EPs only achieve acceptable performance in \textbf{7} adjectives.

\subsection{Effect of PCA}
\label{sec:experiment_pca}
To study how the proposed PCA-based dimension reduction method influences information quality, comparisons were conducted on features produced with different PCA component settings.
All the rest experiments are evaluated from information quality perspective using \textbf{mAP}.
All dataset and PCA setups were the same as in Section~\ref{sec:meth}, except the number of components for PCA. 

\begin{figure}[!t]
    \centering
    \includegraphics[width=\linewidth]{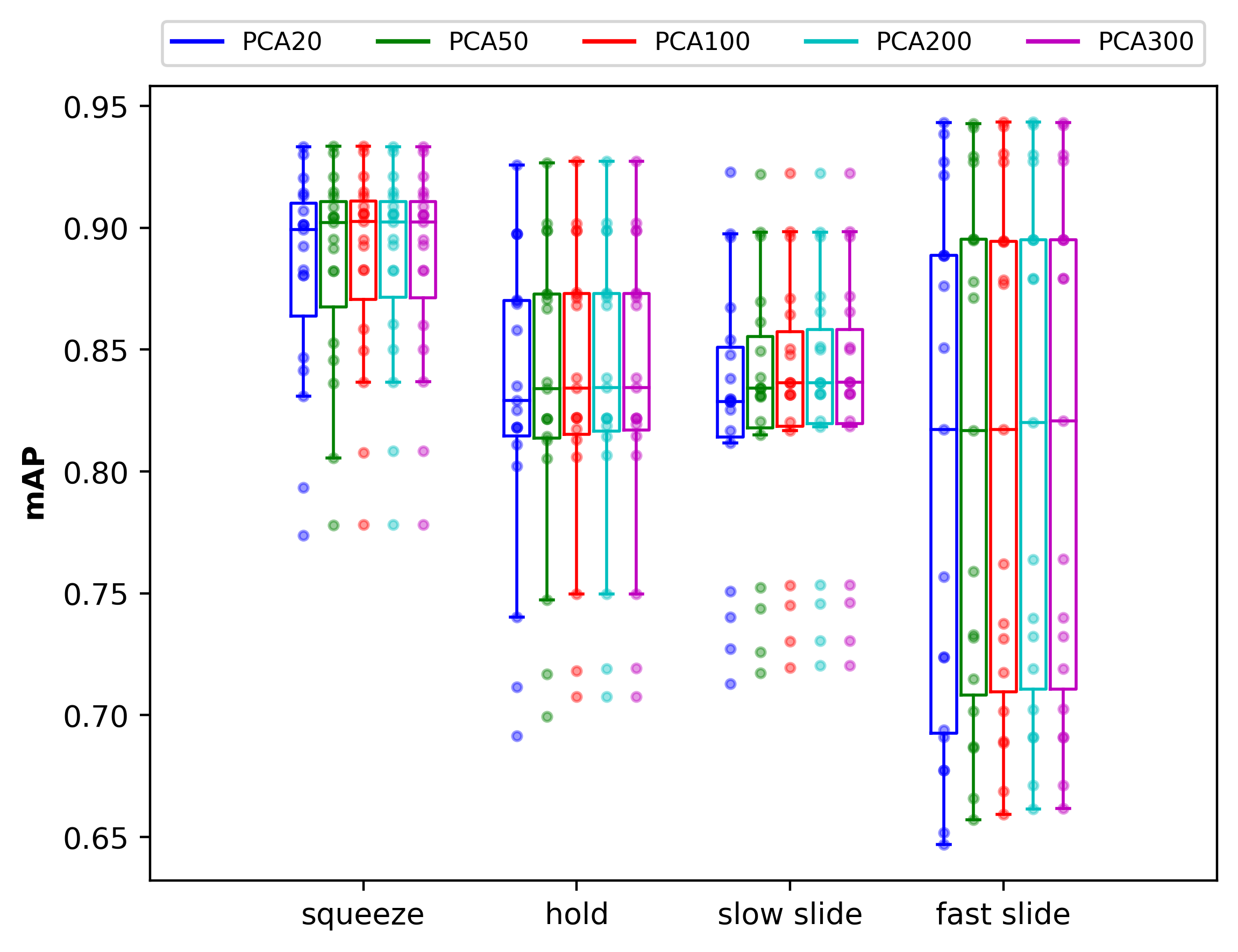}
    \caption{Boxplots of \textbf{mAP} of signals with different numbers of PCA components.}
    \vspace{-1em}
    \label{fig:pca_comp_all_mAP}
\end{figure}

Fig.~\ref{fig:pca_comp_all_mAP} shows the results, from which we can see that signals with more components have slightly better \textbf{mAP} performance up to 200 components. 
PCA with 200 components will be used in the remainder of the experiments due to its lower computational cost (dimensions reduced by more than 99.7\% as shown in Table~\ref{tab:length_of_each_EP}) and no obvious performance loss.

\subsection{Benefit from each tactile sensor}
\label{sec:finger_comp}

 \begin{figure}[!t]
      \centering
      \includegraphics[width=1.0\linewidth]{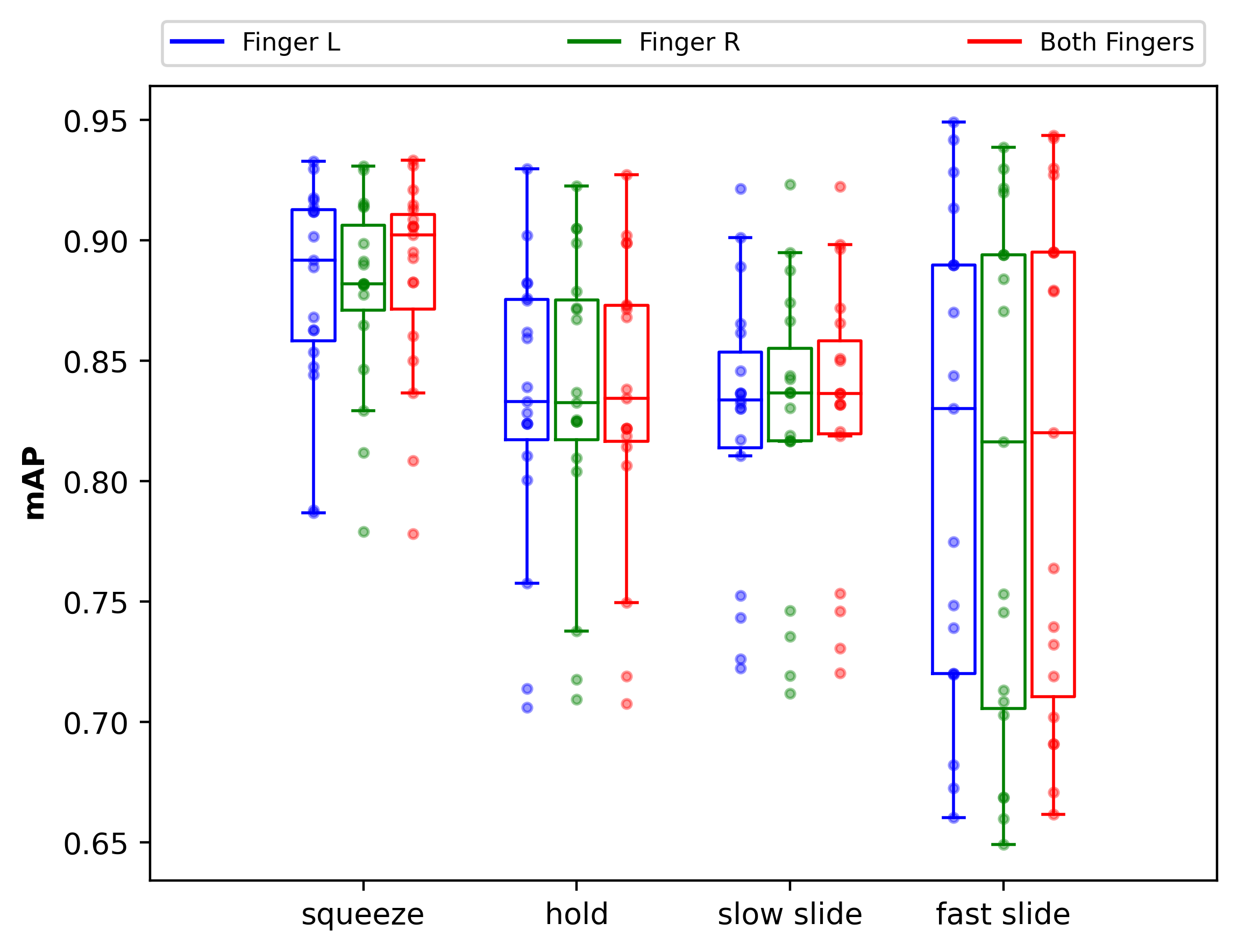}
      \caption{Boxplots of \textbf{mAP} of left finger (Finger L), right finger (Finger R) and both fingers.}
      \label{fig:finger_comp_mAP}
  \end{figure}
  
   \begin{table}[!t]
        \centering
        \begin{tabular}{c|c|c|c|c}
        \hline
             & Squeeze & Hold & Slow Slide & Fast Slide \\
        \hline
            \textbf{Finger L} & 0.881 & 0.833 & 0.826 & 0.810\\
            \textbf{Finger R} & 0.879 & 0.835 & 0.824 & 0.802\\
            \textbf{Both Fingers} & 0.886 & 0.835 & 0.829 & 0.809\\
        \hline
        \end{tabular}
        \caption{Average \textbf{mAP} of different fingers w.r.t. EPs}
        \vspace{-1em}
        \label{tab:finger_avg_f1_scores}
    \end{table}

As introduced in Section~\ref{sec:dataset}, the dataset was collected using a pair of BioTac sensors. This section evaluates how each tactile sensor contributes to the performance in inferring object adjectives. All setups were the same as in Section~\ref{sec:meth} except the signals to which PCA was applied. Fig.~\ref{fig:finger_comp_mAP} shows the \textbf{mAP} results of left finger ($\mathbf{Z}^{200}_{L}$), right finger ($\mathbf{Z}^{200}_{R}$) and both fingers ($\mathbf{Z}^{200}$). We can see that the performance of ``Both Fingers'' is a combination of ``Finger L'' and ``Finger R'' across all EPs. All three cases have very similar performance although with some differences, which can also be observed from the average \textbf{mAP} scores in Table~\ref{tab:finger_avg_f1_scores}. This indicates that one tactile sensor is sufficient for symmetric use cases (symmetric interaction motions on symmetric objects/surfaces).

\begin{figure}[!t]
      \centering
      \includegraphics[width=1.0\linewidth]{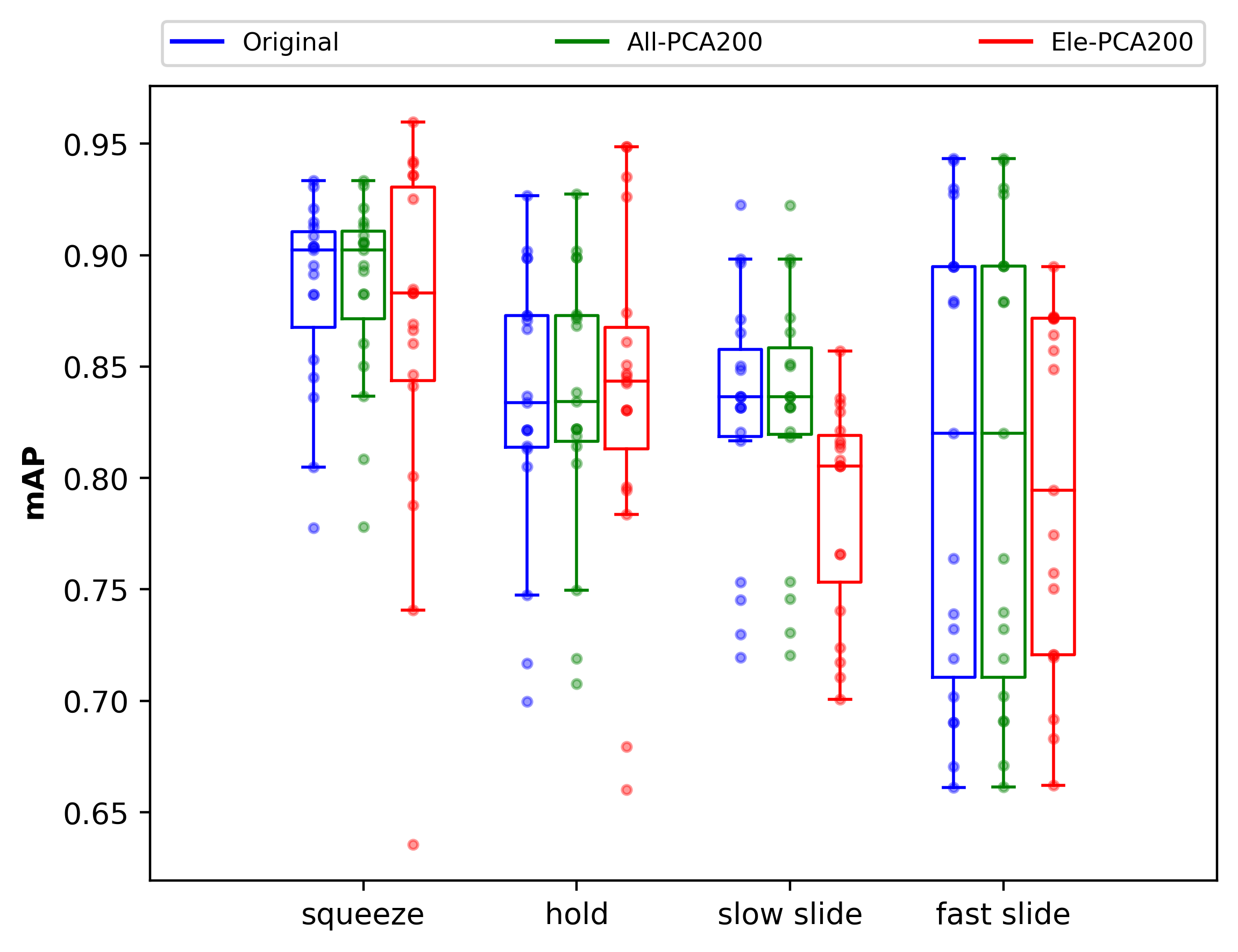}
      \caption{Boxplots of \textbf{mAP} of signals with all four scalars and electrodes (``All-PCA200''), pure electrode signals (``Ele-PCA200''), and original features (``Original'').}
      \label{fig:all_vs_ele_mAP}
  \end{figure}
  
  \begin{figure}[!t]
      \centering
      \includegraphics[width=1.0\linewidth]{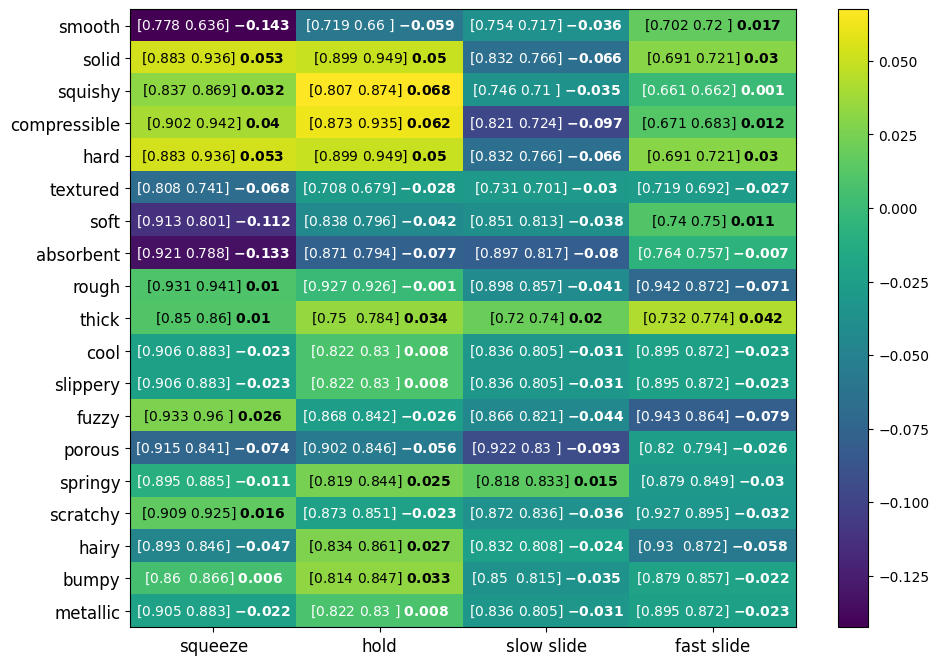}
      \caption{The \textbf{mAP} results of ``All-PCA200'' and ``Ele-PCA200'' across all adjectives and EPs. Their values are shown as  ``[All-PCA200 Ele-PCA200] change'', with the change bolded and their values indicated by the grid colors.}
      \label{fig:all_vs_ele_vis}
      \vspace{-1em}
  \end{figure}

 \begin{figure*}[t]
    \centering
    \includegraphics[width=0.75\linewidth]{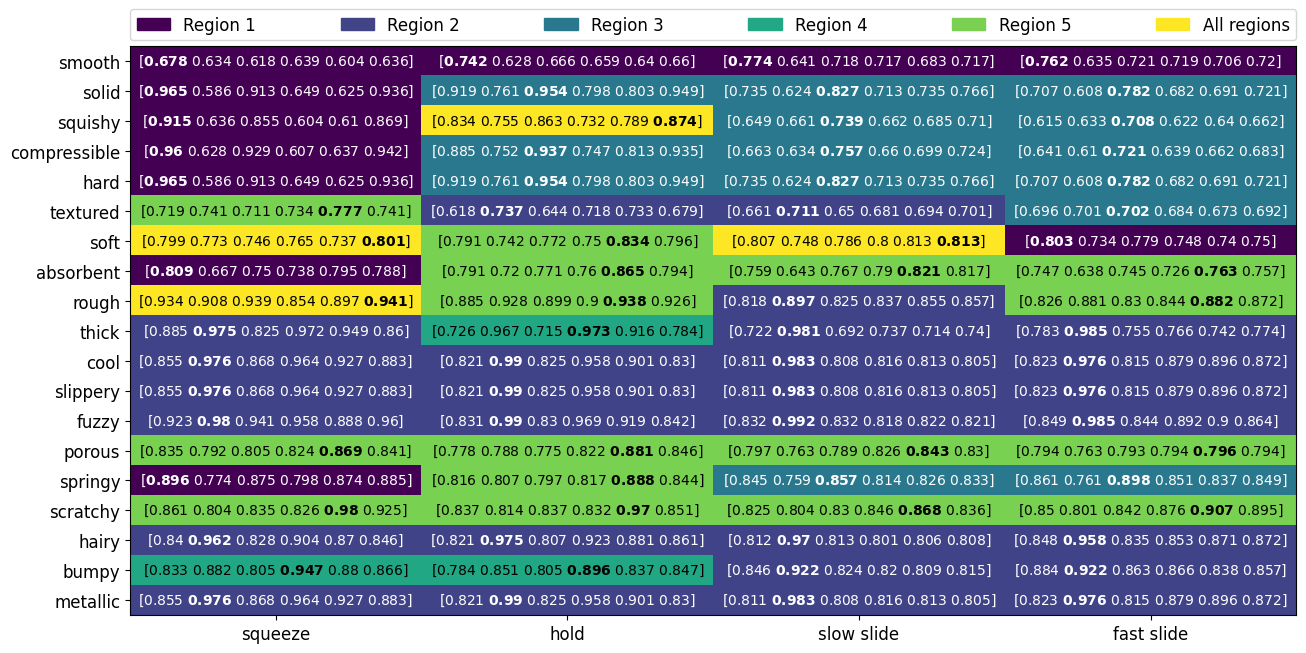}
    \caption{The \textbf{mAP} scores of different regions. Their values are shown as ``[Region 1 to 5, All regions]'', with the best ones bolded and their indexes indicated by the grid colors.}
    \vspace{-1em}
    \label{fig:vis_mAP_regions}
\end{figure*}

\begin{figure}[]
      \centering
      \includegraphics[width=0.9\linewidth]{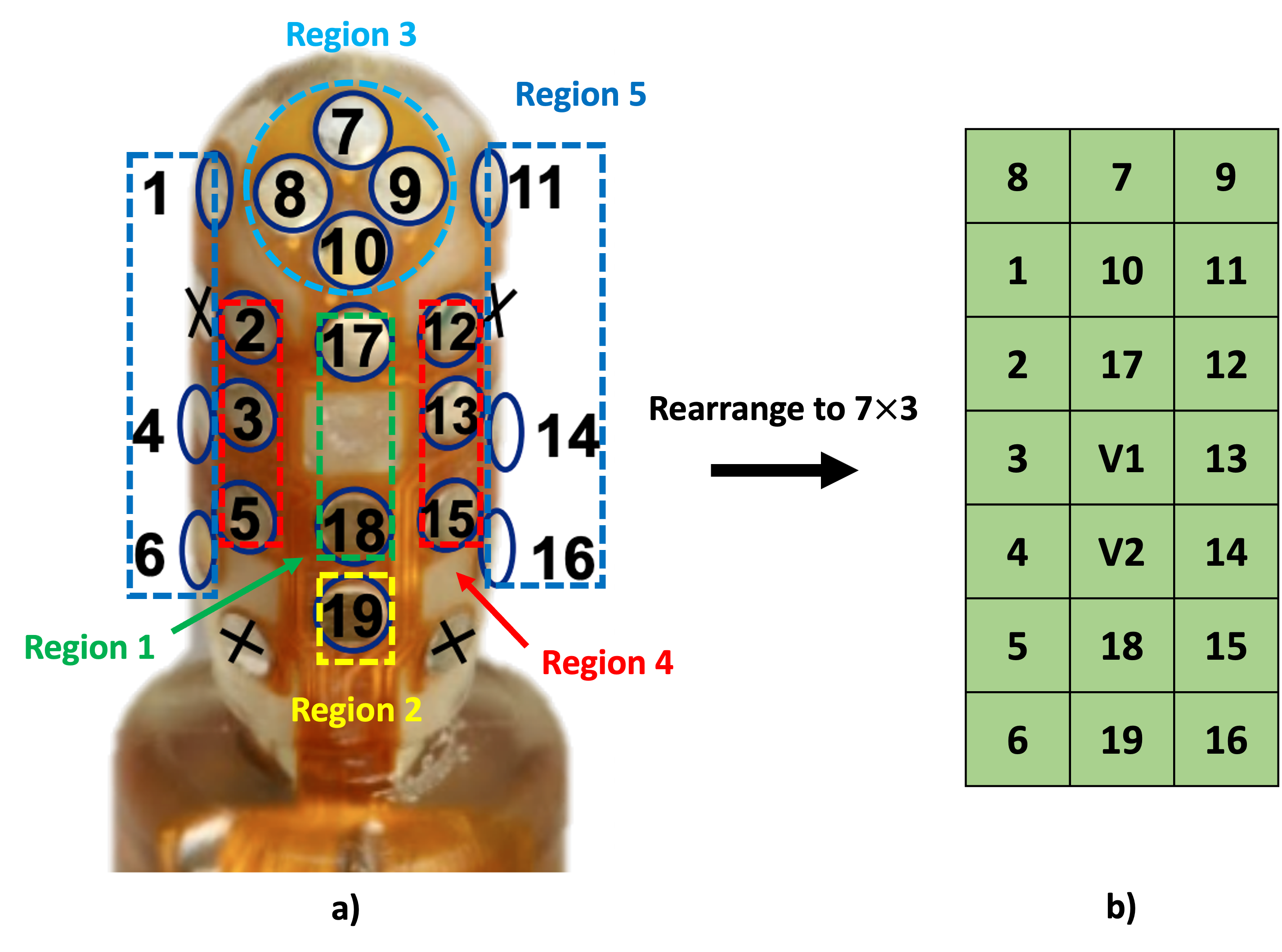}
      \caption{Taxel arrangements on the BioTac sensor. a): The 19 taxels are divided into 5 regions according to their locations on the finger tip. b): Tactile images used for the taxel spatial density study in Section~\ref{sec:spatial_density_results}. The X values are the reference electrodes, the 19 BioTac electrodes are measured relative to these 4 reference electrodes. V1 and V2 are created by taking an average response of the neighboring electrodes: V1 = avg(E17, E18, E12, E2, E13, E3) and V2 = avg(E17, E18, E15, E5, E13, E3)~\cite{chebotar2016self}.}
      \vspace{-2em}
      \label{fig:taxel_arrangement}
  \end{figure}

\subsection{Benefits from taxels and other signals}
\label{sec:all_vs_ele}

Similar to~\cite{richardson2019improving}, comparisons were conducted on different types of signals, but this work focuses more on electrode versus other signals. All setups were the same as in Section~\ref{sec:meth} except the signals which the PCA was applied to. Fig.~\ref{fig:all_vs_ele_mAP} shows the \textbf{mAP} results of signals with all four scalars and taxels (``All-PCA200'') and pure taxel signals (``Ele-PCA200''). The performance of the raw tactile signals with no PCA dimension reduction was also evaluated as a reference (``Original'').

From Fig.~\ref{fig:all_vs_ele_mAP} we can observe that the performance of ``All-PCA200'' is even slightly better than ``Original'' although with greatly fewer dimensions. 
This indicates that the PCA dimension reduction improves the feature quality.
The overall performance of ``Ele-PCA200'' is comparable to ``All-PCA200'' in most EPs except for ``slow slide'' where it is worse than ``All-PCA200''. This can be observed more clearly in Fig.~\ref{fig:all_vs_ele_vis}, from which we can also see that: all texture-related (``smooth'', ``textured'', ``soft'', ``absorbent'', ``rough'', `slippery'', ``fuzzy'', ``porous'', ``scratchy'', and ``hairy'') or temperature-related (``cool'' and ``metallic'') adjectives have worse performance across at least 3 EPs; the others have better performance across at least 2 EPs. This is unsurprising as the four scalar signals contain vibration and temperature related information which is particularly helpful for those adjectives, but not that helpful for the rest (sometimes distracting, such as for ``solid'', ``squishy'', ``compressible'' and ``hard'').

\subsection{Benefits of different taxels}
\label{sec:region_comp}

To explore how each taxel contributes to inferring object adjectives by providing information with a higher quality, comparisons were made on different taxels by dividing the electrodes into 5 regions shown in Fig.~\ref{fig:taxel_arrangement}a. 
The original electrodes (``All regions'') were also compared as a reference. 
All setups were the same as in Section~\ref{sec:meth} except the signals to which the PCA was applied.
The \textbf{mAP} results are shown in
Fig.~\ref{fig:vis_mAP_regions}, from which we can observe that the original electrodes (``All regions'') only worked the best in four cases (``squishy'' with ``hold'', ``soft'' with ``squeeze'' and ``slow slide'', and ``rough'' with ``squeeze''); the best results of the rest were obtained by individual regions. The surprisingly poor performance of ``All regions'' indicates that not all electrodes contribute to inferring object adjectives. This might be caused by noisy or confusing signals from some regions for some adjectives and EPs.

In particular, ``Region 1 to 3'' cover most of the best results, indicating that these regions contribute the most to inferring adjectives. ``Region 1'' tends to contribute more with ``squeeze'', ``Region 3'' contributes more with the rest of three EPs, while ``Region 2'' contributes across all EPs.
The contribution from ``Region 4'' is very limited (the best in only three cases: ``bumpy'' with ``squeeze'' and ``hold'' EPs, and ``thick'' with ``hold''). ``Region 5'' contributes the most in the cases where contacts on either/both sides of the BioTac sensor are critical for inferring certain adjectives, such as ``porous'', ``scratchy'' and ``absorbent''.

  \begin{figure}[t]
      \centering
      \includegraphics[width=1.0\linewidth]{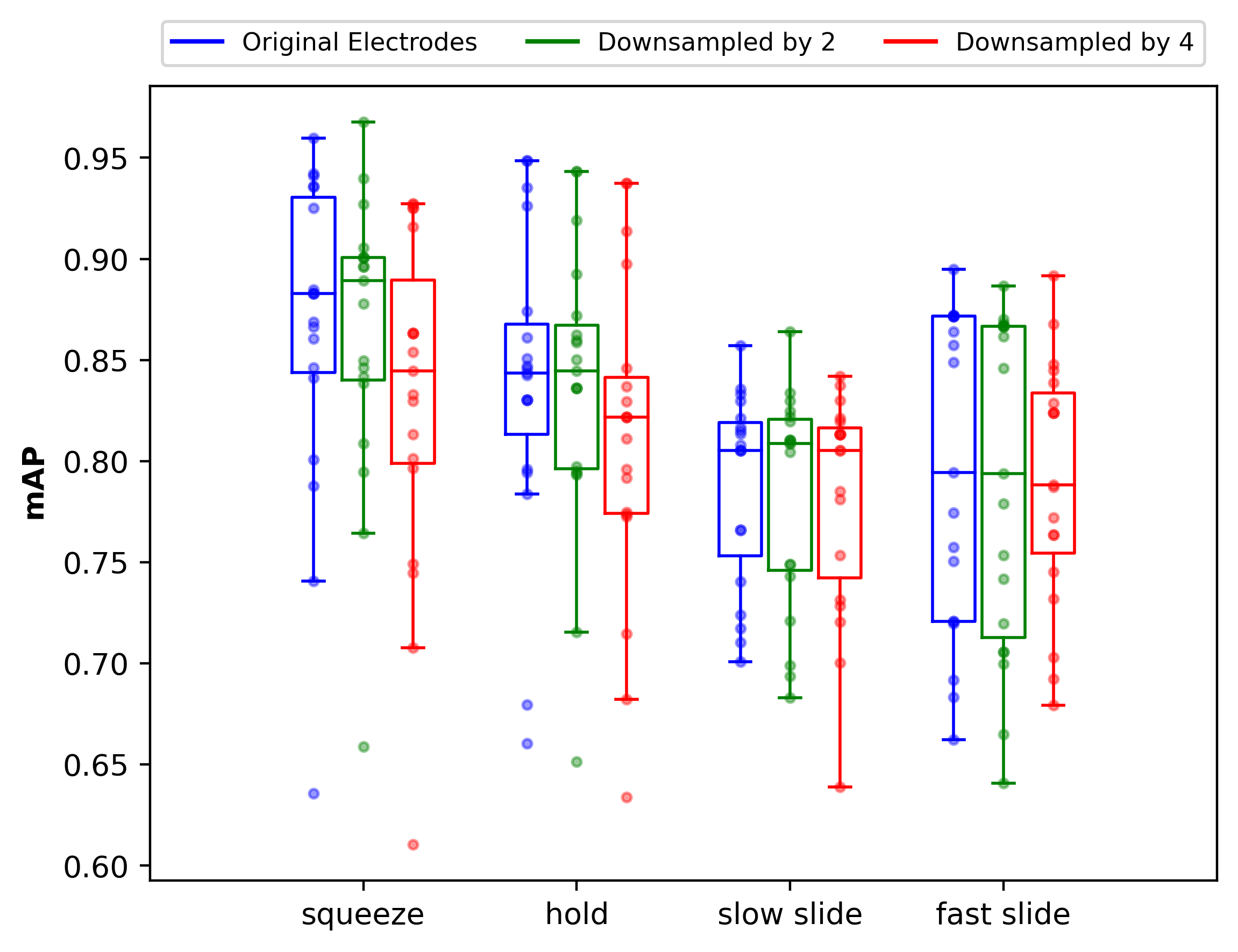}
      \caption{Boxplots of \textbf{mAP} of spatially-downsampled signals and original electrodes.}
      \vspace{-1em}
      \label{fig:ds_mAP}
  \end{figure}
  
    \begin{figure}[t]
      \centering
      \includegraphics[width=1.0\linewidth]{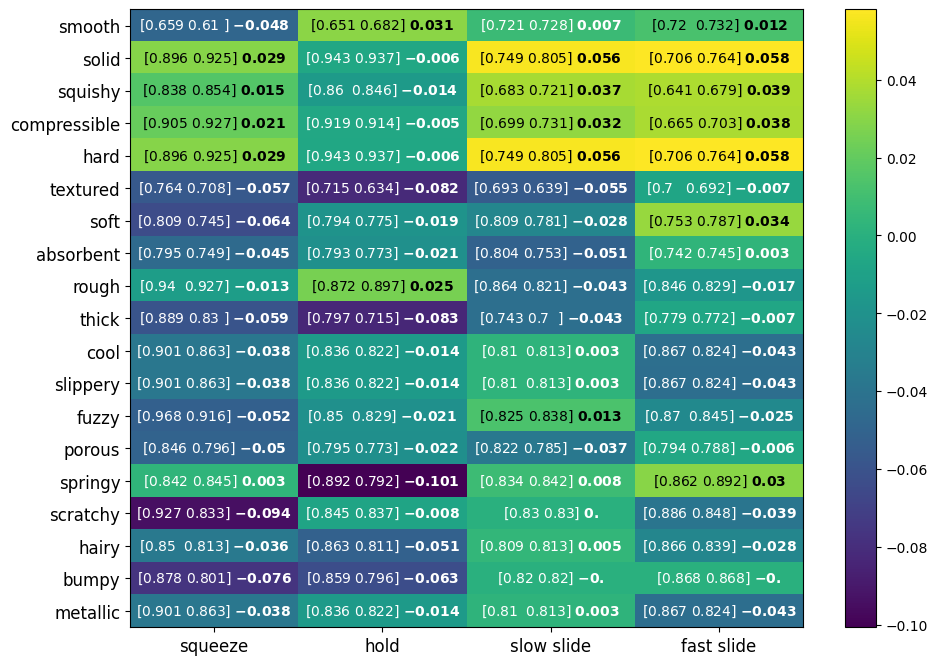}
      \caption{The \textbf{mAP} of the signals spatially downsampled by 2 (\textbf{SDS2}) and 4 (\textbf{SDS4}) across all adjectives and EPs. Their values are shown as  ``[\textbf{SDS2} \textbf{SDS4}] \textbf{SDS2}-\textbf{SDS4}'', with the changes bolded and their values indicated by the grid colors.}
      \vspace{-1em}
      \label{fig:vis_ds_mAP}
    \end{figure}

\subsection{Influence of taxel spatial density}
\label{sec:spatial_density_results}
This section evaluates how the taxel spatial density influences the performance in inferring object adjectives. The 19 electrode measurements from each sensor were first arranged into a $7\times3$ rectangular array using the same method in~\cite{chebotar2016self,richardson2019improving} as shown in Fig.~\ref{fig:taxel_arrangement}b. The study was conducted by comparing the performance of signals spatially-downsampled from 7$\times$3 (original electrodes) to 4$\times$2 (by 2, denoted as \textbf{SDS2}) and 2$\times$1 (by 4, denoted as \textbf{SDS4}). 
All setups were the same as in Section~\ref{sec:meth} except the signals to which the PCA was applied.

Fig.~\ref{fig:ds_mAP} and Fig.~\ref{fig:vis_ds_mAP} show the results. From Fig.~\ref{fig:ds_mAP} we can see that downsampling generally decreases the performance in most cases across all EPs. 
From adjective perspective (Fig.~\ref{fig:vis_ds_mAP}), ``solid'', ``squishy'', ``compressible'' and ``hard'' have obviously better performance with downsampled signals across 3 EPs except ``hold''; 
most of the rest have obvious performance decreases across at least 3 EPs, particularly for those texture-related ones. This indicates that denser taxels are helpful for those texture-related adjectives, but can be distracting to those non-texture-related ones.

\subsection{Influence of measurement temporal density}

\begin{figure*}[]
    \centering
    \includegraphics[width=1.0\linewidth]{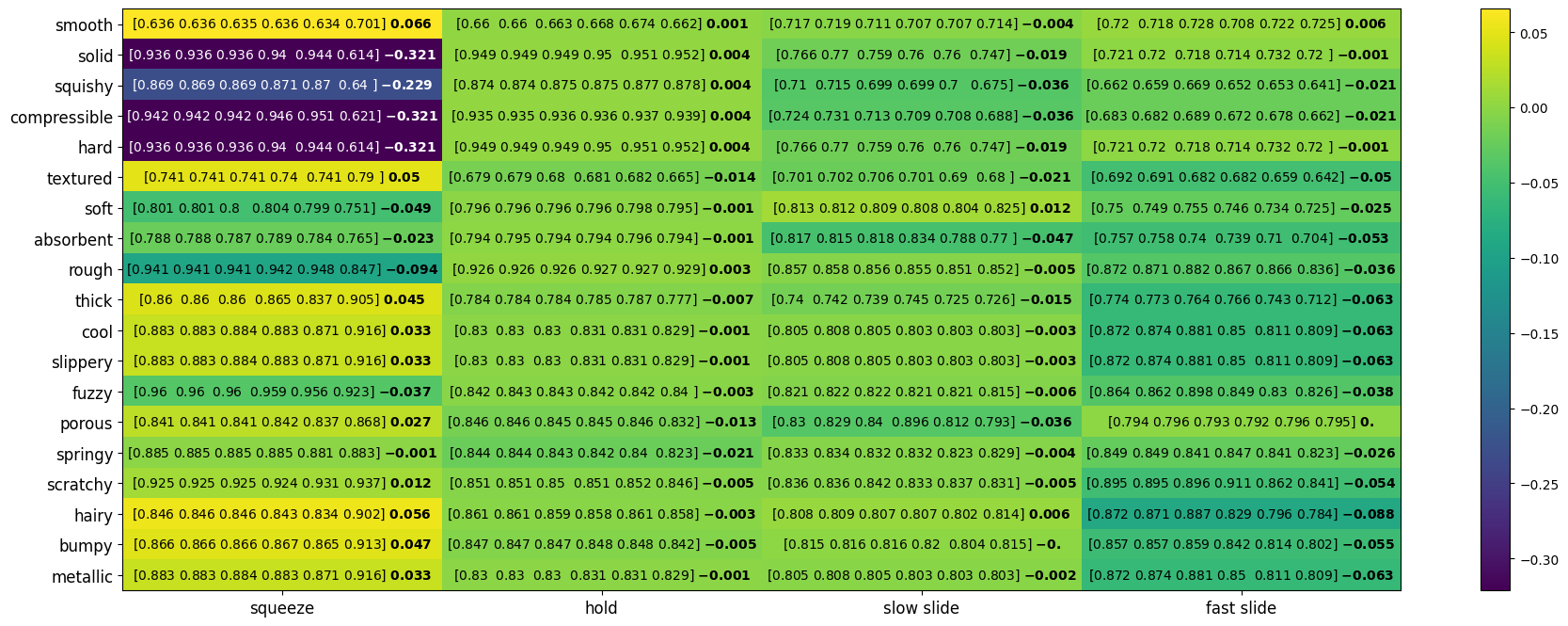}
    \caption{The \textbf{mAP} results of the temporally-downsampled signals across all adjectives and EPs. Their values are shown as  ``[\textbf{TDS1} to \textbf{TDS400}] \textbf{TDS400}-\textbf{TDS1}'', with the changes bolded and their values indicated by the grid colors.}
    \label{fig:vis_temp_ds_mAP}
    \vspace{-1em}
\end{figure*}

\begin{figure}[]
    \centering
    \includegraphics[width=1.0\linewidth]{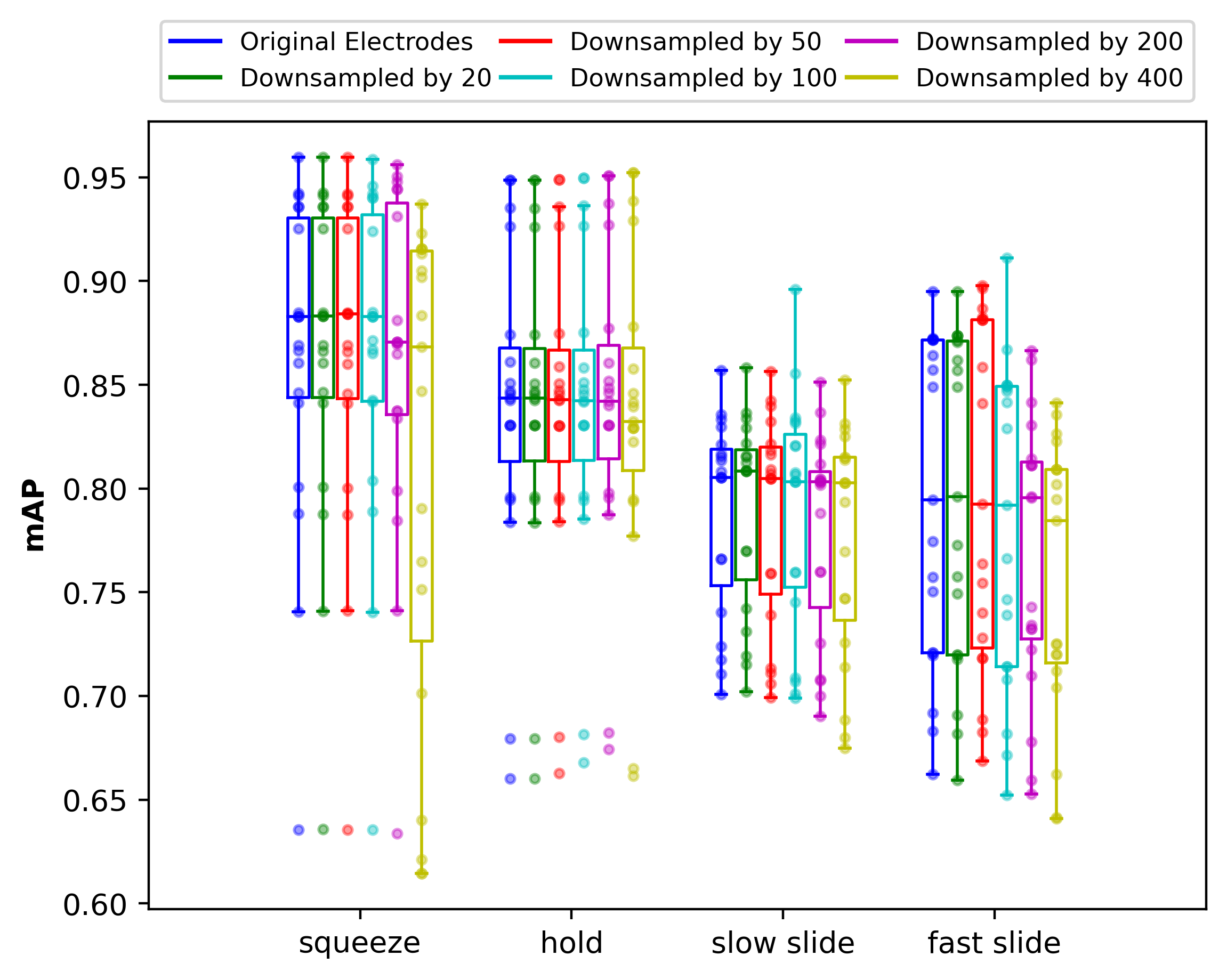}
    \caption{Boxplots of \textbf{mAP} of temporally-downsampled signals and original electrodes.}
    \vspace{-1em}
    \label{fig:temp_ds_mAP}
\end{figure}

The study of measurement temporal density was conducted by comparing signals temporally downsampled by factors of 20, 50, 100, 200, and 400 (denoted as \textbf{TDS20} to \textbf{TDS400}).
The performance of the original electrodes was also compared as a reference (denoted as \textbf{TDS1}).
All setups were the same as in Section~\ref{sec:meth} except the signals to which the PCA was applied. Considering the significantly reduced number of frames in the downsampled sequences, the numbers of PCA components were set to 100 for \textbf{TDS200} and 50 for \textbf{TDS400}, which maintained $>99\%$ of the input's variance.

Table~\ref{tab:temp_ds_avg_mAP} shows the average \textbf{mAP} results, from which we can observe that signals collected by ``squeeze'' have obvious performance degradation ($\geqslant0.01$) from \textbf{TDS400}, ``slow slide'' from \textbf{TDS200}, and ``fast slide'' from \textbf{TDS100}; while ``hold'' has no obvious performance changes. This trend can also be observed in Fig.~\ref{fig:temp_ds_mAP}. When paying attention to the numbers of frames in the downsampled sequences of each EP in Table~\ref{tab:temp_ds_avg_mAP}, we can also observe that the performance starts decreasing at around 10 for ``squeeze'' (12) and ``slow slide'' (9), and 18 for ``fast slide''. There is no obvious performance difference for the cases with smaller downsampling factors. All these observations indicate that much smaller frame rates can satisfy the demand of inferring object adjectives, although with differences for some adjectives and EPs. For those exploratory procedures with the possibility of obtaining more information, more frames are required to keep the information.

  \begin{table}[t]
    \centering
    \setlength\tabcolsep{4.5pt}
    \begin{tabular}{c|c|c|c|c}
    \hline
          & Squeeze & Hold & Slow Slide & Fast Slide \\
         \hline
         \textbf{TDS1} & 0.864 (4740) & 0.838 (2772)  & 0.788 (1784)  & 0.795 (1714) \\
         \textbf{TDS20}& 0.864 (237) & 0.838 (139) & 0.790 (90) & 0.794 (86) \\
         \textbf{TDS50}& 0.864 (95) & 0.838 (56) & 0.787 (36) & 0.798 (35) \\
         \textbf{TDS100}& 0.865 (48) & 0.839 (28) & 0.789 (18) & 0.783 (18) \\
         \textbf{TDS200}& 0.861 (24) & 0.840 (14)  & 0.779 (9)  & 0.769 (9) \\
         \textbf{TDS400}& 0.812 (12) & 0.835 (7)  & 0.775 (5)  & 0.757 (5) \\
        \hline
    \end{tabular}
    \caption{The average \textbf{mAP} results of temporally-downsampled signals, with the numbers of frames in downsampled sequences of each EP included in the brackets.}
    \vspace{-1em}
    \label{tab:temp_ds_avg_mAP}
\end{table}

From Fig.~\ref{fig:vis_temp_ds_mAP} we can also see that the performance changes of ``solid'', ``squishy'', ``compressible'' and ``hard'' with ``squeeze'' are much more significant ($\geqslant0.1$) than the others. This indicates that the contributions from the temporal information in these signals are critical for inferring stiffness-related adjectives, and the minimum numbers of frames are around 24 to keep the information.
   
\section{Conclusions and Discussions}

This paper has presented a comprehensive empirical study on how parallel finger-tip tactile contributes to inferring object adjectives using the PHAC-2 dataset. The study was mainly conducted to explore the influence of taxel spatial density, measurement temporal density, and taxel distribution. Experimental results showed that \textbf{one tactile sensor is sufficient for symmetric use cases} (symmetric interaction motions on symmetric objects/surfaces) such as most of the data collected in the PHAC-2 dataset. The two tactile sensors on the PR2 robot have similar performance individually or combined.

Both spatial and temporal information from taxels contributes to inferring object adjectives. \textbf{Dense taxels are beneficial for texture-related adjectives, but can be distracting to non-texture-related ones}. The contributions from the temporal information are critical for inferring stiffness-related adjectives such as ``solid'', ``squishy'', ``compressible'' and ``hard''. \textbf{The frame-rate of the BioTac sensor is more than sufficient for inferring object adjectives}. Sensors with much smaller frame rates can satisfy the demand. 

The vibration and temperature related signals from the BioTac sensor have significant contributions to the inferring of textured-related or temperature-related adjectives, \textbf{but can also be distracting for some adjectives} such as stiffness-related features. In terms of taxel distribution, \textbf{not all taxels benefit the inferring of object adjectives}: the electrodes on the belly of the sensor make the largest contribution to inferring the most object adjectives, while the electrodes on both sides of the sensor are critical for inferring the adjectives such as ``porous'', ``scratchy'' and ``absorbent'', where deep contact (sensor mostly wrapped by the contact surface) between the sensor and object is needed.

\section*{ACKNOWLEDGMENT}
The authors acknowledge continued support from the Queensland University of Technology (QUT) through the Centre for Robotics, and thank the support and help from Dr Ben Richardson for the use of the PHAC-2 dataset.
Support from the GentleMAN (RCN 299757) is also greatly acknowledged.
Computational resources and services used in this work were provided by the eResearch Office, Queensland University of Technology, Brisbane, Australia.

\bibliographystyle{IEEEtran}
\bibliography{fangyi}

\begin{thebibliography}{10}
\providecommand{\url}[1]{#1}
\csname url@samestyle\endcsname
\providecommand{\newblock}{\relax}
\providecommand{\bibinfo}[2]{#2}
\providecommand{\BIBentrySTDinterwordspacing}{\spaceskip=0pt\relax}
\providecommand{\BIBentryALTinterwordstretchfactor}{4}
\providecommand{\BIBentryALTinterwordspacing}{\spaceskip=\fontdimen2\font plus
\BIBentryALTinterwordstretchfactor\fontdimen3\font minus
  \fontdimen4\font\relax}
\providecommand{\BIBforeignlanguage}[2]{{%
\expandafter\ifx\csname l@#1\endcsname\relax
\typeout{** WARNING: IEEEtran.bst: No hyphenation pattern has been}%
\typeout{** loaded for the language `#1'. Using the pattern for}%
\typeout{** the default language instead.}%
\else
\language=\csname l@#1\endcsname
\fi
#2}}
\providecommand{\BIBdecl}{\relax}
\BIBdecl

\bibitem{chu2015robotic}
V.~Chu, I.~McMahon, L.~Riano, C.~G. McDonald, Q.~He, J.~M. Perez-Tejada,
  M.~Arrigo, T.~Darrell, and K.~J. Kuchenbecker, ``Robotic learning of haptic
  adjectives through physical interaction,'' \emph{Robotics and Autonomous
  Systems}, vol.~63, pp. 279--292, 2015.

\bibitem{gao2016deep}
Y.~Gao, L.~A. Hendricks, K.~J. Kuchenbecker, and T.~Darrell, ``Deep learning
  for tactile understanding from visual and haptic data,'' in \emph{2016 IEEE
  International Conference on Robotics and Automation (ICRA)}.\hskip 1em plus
  0.5em minus 0.4em\relax IEEE, 2016, pp. 536--543.

\bibitem{richardson2019improving}
B.~A. Richardson and K.~J. Kuchenbecker, ``Improving haptic adjective
  recognition with unsupervised feature learning,'' in \emph{2019 International
  Conference on Robotics and Automation (ICRA)}.\hskip 1em plus 0.5em minus
  0.4em\relax IEEE, 2019, pp. 3804--3810.

\bibitem{richardson2020learning}
------, ``Learning to predict perceptual distributions of haptic adjectives,''
  \emph{Frontiers in Neurorobotics}, vol.~13, p. 116, 2020.

\bibitem{liu2017recent}
H.~Liu, Y.~Wu, F.~Sun, and D.~Guo, ``Recent progress on tactile object
  recognition,'' \emph{International Journal of Advanced Robotic Systems},
  vol.~14, no.~4, p. 1729881417717056, 2017.

\bibitem{navarro2022visuo}
N.~Navarro-Guerrero, S.~Toprak, J.~Josifovski, and L.~Jamone, ``Visuo-haptic
  object perception for robots: An overview,'' \emph{arXiv preprint
  arXiv:2203.11544}, 2022.

\bibitem{liu2016extreme}
H.~Liu, J.~Qin, F.~Sun, and D.~Guo, ``Extreme kernel sparse learning for
  tactile object recognition,'' \emph{IEEE transactions on cybernetics},
  vol.~47, no.~12, pp. 4509--4520, 2016.

\bibitem{liu2017multi}
H.~Liu, Y.~Wu, F.~Sun, D.~Guo, and B.~Fang, ``Multi-label tactile property
  analysis,'' in \emph{2017 IEEE International Conference on Robotics and
  Automation (ICRA)}.\hskip 1em plus 0.5em minus 0.4em\relax IEEE, 2017, pp.
  366--371.

\bibitem{liu2017structured}
H.~Liu, F.~Sun, D.~Guo, B.~Fang, and Z.~Peng, ``Structured output-associated
  dictionary learning for haptic understanding,'' \emph{IEEE Transactions on
  Systems, Man, and Cybernetics: Systems}, vol.~47, no.~7, pp. 1564--1574,
  2017.

\bibitem{abderrahmane2018visuo}
Z.~Abderrahmane, G.~Ganesh, A.~Crosnier, and A.~Cherubini, ``Visuo-tactile
  recognition of daily-life objects never seen or touched before,'' in
  \emph{2018 15th International Conference on Control, Automation, Robotics and
  Vision (ICARCV)}.\hskip 1em plus 0.5em minus 0.4em\relax IEEE, 2018, pp.
  1765--1770.

\bibitem{zhu2004recall}
M.~Zhu, ``Recall, precision and average precision,'' \emph{Department of
  Statistics and Actuarial Science}, vol.~2, 2004.

\bibitem{taha2015metrics}
A.~A. Taha and A.~Hanbury, ``Metrics for evaluating 3d medical image
  segmentation: analysis, selection, and tool,'' \emph{BMC medical imaging},
  vol.~15, no.~1, pp. 1--28, 2015.

\bibitem{jolliffe2016principal}
I.~T. Jolliffe and J.~Cadima, ``Principal component analysis: a review and
  recent developments,'' \emph{Philosophical transactions of the royal society
  A: Mathematical, Physical and Engineering Sciences}, vol. 374, no. 2065, p.
  20150202, 2016.

\bibitem{kecman2005support}
V.~Kecman, ``Support vector machines--an introduction,'' in \emph{Support
  vector machines: theory and applications}.\hskip 1em plus 0.5em minus
  0.4em\relax Springer, 2005, pp. 1--47.

\bibitem{burges1998tutorial}
C.~J. Burges, ``A tutorial on support vector machines for pattern
  recognition,'' \emph{Data mining and knowledge discovery}, vol.~2, no.~2, pp.
  121--167, 1998.

\bibitem{chebotar2016self}
Y.~Chebotar, K.~Hausman, Z.~Su, G.~S. Sukhatme, and S.~Schaal,
  ``Self-supervised regrasping using spatio-temporal tactile features and
  reinforcement learning,'' in \emph{2016 IEEE/RSJ International Conference on
  Intelligent Robots and Systems (IROS)}.\hskip 1em plus 0.5em minus
  0.4em\relax IEEE, 2016, pp. 1960--1966.

\end{thebibliography}

\end{document}